# CE301 Final Report

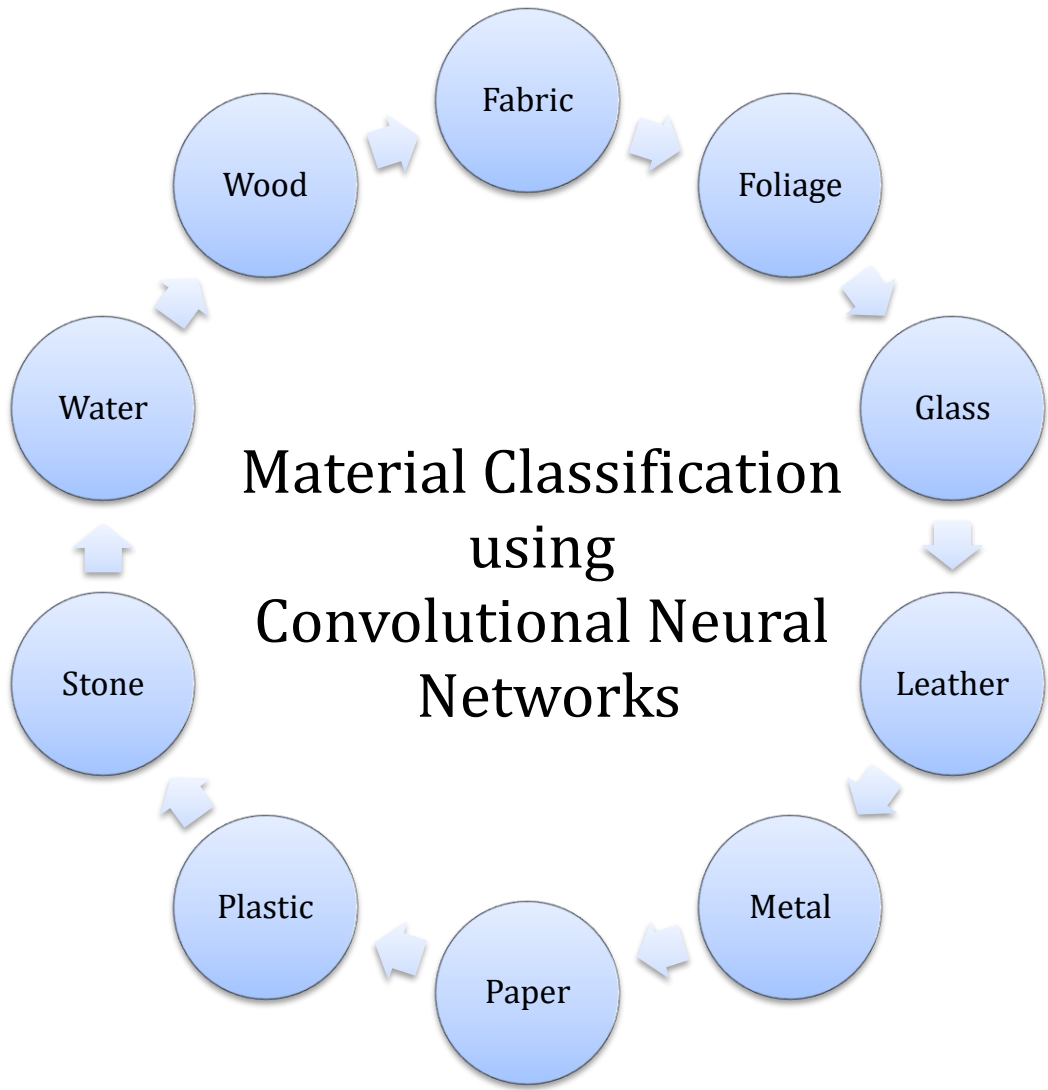

Material Classification using Convolutional Neural Networks

**Student:** Anca Sticlaru

**Registration Number**: 1301000

**Supervisors:** Klaus McDonald-Maier / Shoaib Ehsan

**Second Assessor:** Nikolaos Thomos

**Degree Course:** BEng Computers with Electronics

# Acknowledgements

I would like to acknowledge my supervisors Prof. Klaus McDonald-Maier and Dr. Shoaib Ehsan for providing me the necessary guidance and valuable support throughout this research project. I value the assistance of Grigorios Kalliatakis. Learning from their knowledge helped me to become passionate about my research topic.

# Abstract

The recognition and classification of the diversity of materials that exist in the environment around us are a key visual competence that computer vision systems focus on in recent years. Understanding the identification of materials in distinct images involves a deep process that has made usage of the recent progress in neural networks which has brought the potential to train architectures to extract features for this challenging task. This project uses state-of-the-art Convolutional Neural Network (CNN) techniques and Support Vector Machine (SVM) classifiers in order to classify materials and analyze the results. Building on various widely used material databases collected, a selection of CNN architectures is evaluated to understand which is the best approach to extract features in order to achieve outstanding results for the task. The results gathered over four material datasets and nine CNNs outline that the best overall performance of a CNN using a linear SVM can achieve up to ~92.5% mean average precision, while applying a new relevant direction in computer vision, transfer learning. By limiting the amount of information extracted from the layer before the last fully connected layer, transfer learning aims at analyzing the contribution of shading information and reflectance to identify which main characteristics decide the material category the image belongs to. In addition to the main topic of my project, the evaluation of the nine different CNN architectures, it is questioned if, by using the transfer learning instead of extracting the information from the last convolutional layer, the total accuracy of the system created improves. The results of the comparison emphasize the fact that the accuracy and performance of the system improves, especially in the datasets which consist of a large number of images.

# Contents

# Chapter 1 Supporting Literature

## 1.1 Introduction

The first intention of trying to "understand the scene" is one of the base ideas in computer vision [1] that lead to a continuous increase in the need to apprehend the high-level context in images regarding object recognition and image classification. By becoming a fundamental visual expertise that Computer Vision systems require, the field has rapidly grown. Images have become ubiquitous in a variety of fields as so many people and systems extract vast amounts of information from imagery. Information that can be vital in areas such as robotics, hospitals, self-driving cars, surveillance or building 3D representations of objects. While each of the above-mentioned applications differs by numerous factors, they share the common process of correctly annotating an image with one or a probability of labels that correlates to a series of classes or categories. This procedure is known as image classification and, combined with machine learning, it has become an important research topic in the field, on account of the focus on the understanding of what an image is representative of. The complex process of identifying the type of materials in diverse tasks linked to image-based scene perspectives has taken advantage of the combination of machine learning techniques applied to the up-to-date development of neural networks. This outlines the challenging problem of material classification due to the variety of the definite features of materials. The state-of-the-art solutions rely massively on the attention that Computer Vision systems have received, which led to a series of algorithms being developed and images being collected in datasets.

People are able to recognize the environment they are in as well as the various objects in their everyday life no matter the influence on the item's features or if their view is obstructed, as this is one of the very first skills we learn from the moment we are born. Computers, on the other hand, require effort and powerful computation and complex algorithms to attempt to recognize correctly patterns and regions where a possible object might be. Object detection and recognition are two main ways that have been implemented over multiple decades that are at the center of Computer Vision systems at the moment. These approaches are presented with challenges such as scale, occlusion, view point, illumination or background clutter, all issues that have been attempted as research topics that provided functionality that led to the introduction of Neural Networks and Convolutional Neural Networks (CNN) [2]. The newly added functionality is composed of distinct types of layers that consist of many parameters that are able to figure out the features present in a given image. These architectures have since been built on and a more complex structure with hidden non-linear layers between the input and output layers of a CNN has been identified as Deep Convolutional Neural Network (DCNN) [3, 4]. The advancement in the computational speed of computers and the amount of data available has allowed deep learning to increase the overall performance under supervised learning conditions. Results have been shown to be better when there is more data available used by bigger models on a faster system. Favoring the latest research, scientists brought into attention two new topics: object localization [5] and semantic segmentation [6, 7].

The first step of approaching the demanding issue of image classification is by looking at the available data. Material datasets started to be gathered just over a decade ago as only a few number of computer vision systems targeted material recognition as a research topic. Analyzing Columbia-Utrecht Reflectance and Texture Database CUReT [8], the first published dataset, it can be observed that the environment conditions are limited. Although it attempts its best at simulating the real-world settings through having under 205 viewings and an extensive amount illumination directions and having a performance greater than 95%, the fact that it uses synthetic data and a very low number of images per category makes the results inaccurate and limited in capturing the complexity of the materials found in the real-world [9]. This conclusion was emphasized in a low accuracy of just 23% obtained when the same approach was used on the Flickr Materials Dataset (FMD) produced by Sharan et al [10]. The main noticeable difference was the fact that FMD database was using real-world images and increased the number of images per category to 100 compared to CUReT dataset that only had 61 images. The OpenSurfaces dataset [11, 12] is formed of real-world images and achieves a 34.5% accuracy. This dataset takes a step further and focuses on objects from consumer photographs and introduces a larger database with 53 classes and 25000 scenes with more than 110000 segmented materials. An important achievement in the field was presented shortly by Liu et al [13] when the issue of recognizing the type of material from its own features was tackled and a better accuracy of 45% was found. This offered the start for extended research into the areas of shape-based object recognition or texture recognition by developing an algorithm that successfully finds the object and extracts features like its shape, color, texture and reflectance. The algorithm uses a technique called Bag-of-Visual-Words (BoVW) [14] where the words are defined by the features extracted and the bag-of-words is represented by the picture. Cimpoi et al [15] have later released the Describable Textures Dataset (DTD) where for the first time an Improved Fisher Vector (IFV) and a CNN architectures have been combined to outline that together the two approaches had better found the key characteristics of objects by outperforming previous work by 10% and becoming the state-of-the-art. With deep learning achieving better results, Gibert et al [16] proved that by using DCNNs specifically trained for the task of material recognition on railway tracks, the system produces an output of 93.35% efficiency and that such systems continue to produce the state-of-the-art results for image classification [17, 18] and object detection [7]. In the same year, Bell et al [19], has achieved results of 85.2% mean accuracy using CNNs and a brand-new dataset called Materials in Context Database (MINC) that has around 2500 images per category and has a total of 23 classes.

In Computer Vision systems, datasets are divided into two main categories: a training dataset used for training the algorithm learn to perform its desired task and a testing dataset that the algorithm is tested on. The percentages that one divides them by affects the general pipeline and it is the first step when trying to solve this challenging task. The work that is being offered within this report, focuses on a 50% split across all datasets tested. Furthermore, each training and testing dataset is further split into negative and positive structures in order to make sure that the algorithm learns what it should be looking for in an image as well as what not to look for. At the end of a test run, one needs to look at all accuracy counts to be able to come up with a total analysis of the pipeline created. This need emphasizes the fact that the system learns how to better optimize itself. Using previous research work as a base, a comparison of several types of CNN architectures is made. To have as conclusive results a complex and full analysis as well as making sure that the learning process is beneficial for the entire pipeline, and particularly for the feature extractors, four widely used main datasets have been selected for training and testing purposes.

Transfer learning is a promising technique that improves the model learning and is used for object recognition when there are very few training samples and when the need for labeling data needs to be reduced. This approach can be compared to being an augmentation as the information is transferred between layers or classes. In the work by Lim et al [20], it is used for object detection where the developed framework borrows examples from categories in order to learn how to transform those examples to become more similar to the instances from the targeted class. Taking advantage of the information already available in other classes leads to better results for object recognition in [21, 22] which try to adapt the classifiers from the pipeline to overcome the fact that there are not enough training images. By receiving lots of attention because it shows an improvement on the overall results, this approach has become the latest method that Wieschollek and Lensch [23] used to test on FMD dataset and on a new dataset called Google Material Dataset with around 10000 images divided into 10 material categories. Their work has been able to outperform earlier results found on FMD when natural images are used.

## 1.2 Research Contribution Aim

This project builds on countless state-of-the-art methods that have been developed over the last years to compare and analyze the best accurate feature extractor when presented with material classification and provides a state-of-the-art evaluation on the best concepts and effective ideas for the task of classifying materials.

My research aim is in developing a better understanding of the best adapted techniques to the task of material classification, performing a thorough evaluation of a range of nine CNN architectures including latest models as well as comparing and analyzing the factors that lead to structuring a good dataset for material classification and how it affects the overall system. The focus for this goal is to find the high-level categories in the input images in at least seven common categories - ceramic, fabric, glass, metal, paper, plastic and wood - across all databases. In addition to this topic, it is examined whether by applying transfer learning between the layers of each CNN architecture, the system's accuracy will produce any noticeable effects on the gathered results. The work presented in [24] emphasizes similar method of comparing CNN architectures. In comparison with research contribution reported in [24] which focuses on recognizing various object categories in an image and using data augmentations, current research aims at providing a complex evaluation of the latest methods on material datasets in computer vision.

The contribution for this research topic is structured in running a large number of experimental tests on countless training and testing datasets to further expand the understanding to which technique is best for material classification and observe whether transfer learning manages to improve results on the current chosen databases. As the current system will return a ranked sequence of scores and documents, there is a need for a good unit to compute the precision-recall curve and results. A good classifier can rank material images at the top of the returned list. The main performance unit regarding precision-call is called average precision (AP) and it is going to be used on three of the four selected datasets. Compared to just computing and analyzing the precision-recall, the average precision offers a simpler way since it is returning a single number that outlines the performance of the classifier and is computed as:

$$AP = \sum_{k=1}^{n} P(k)\, \Delta r(k)$$

where the sum is the precision at a cutoff of k images multiplied by the change in the recall. If any slight change it is made in the ranking, it does not affect the score very much, which makes this unit stable and preferred by various researchers. This unit will not be able to be used on the last dataset since each category has a distinct number of photographs per class and the outcomes will be interpreted under specific conditions.

This report starts by introducing the pipeline used for the system, including the several types of visual models and the basic linear classifier used for the supervised learning task. Following this, a series of approaches are described to define the feature extractor hierarchy that is concluded with the results and their analysis. Finally, future directions related to this work are discussed.

# Chapter 2 Project description

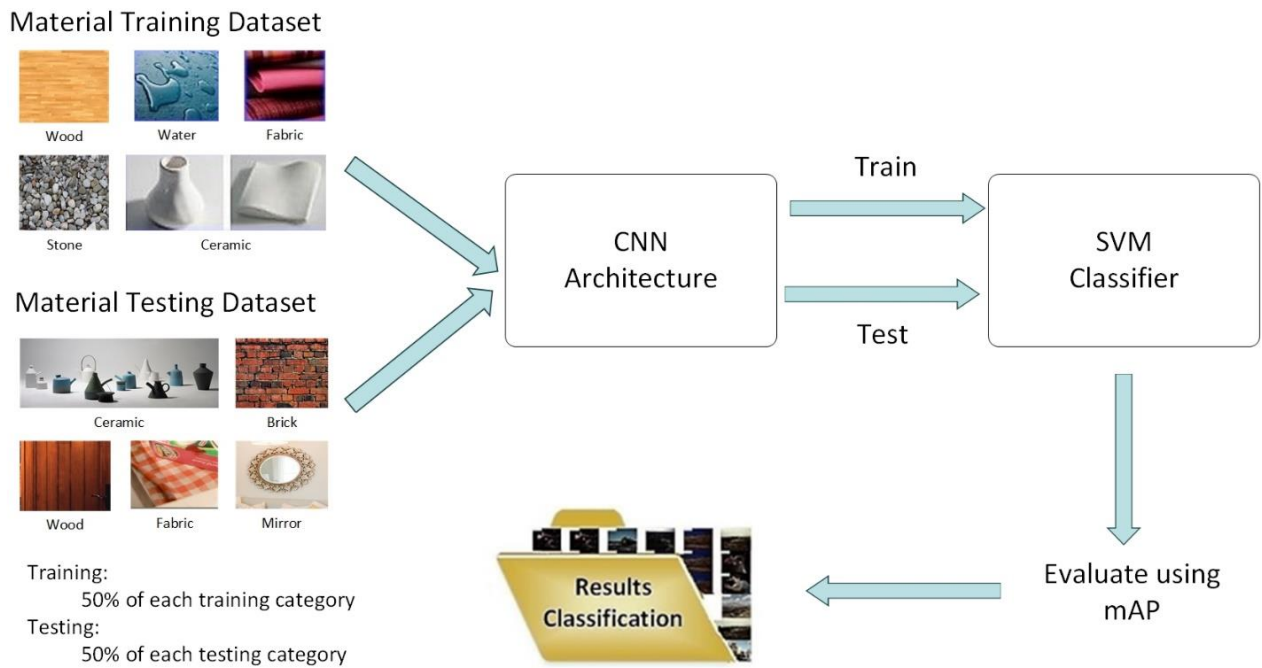

Figure 1 – Material Classification Pipeline

## 2.1 Project pipeline

The project's pipeline presented in Figure 1 consists of four main categories of elements. The input component refers to material categories extracted from the four datasets chosen for evaluation purposes. Each database has been split equally for the training and testing phases of the process. Following this step, nine CNN architectures have been selected and adapted to the Caffe environment in MATLAB [24]. The feature extractors are then passed over to the linear Support Vector Machine which is similar to the one published by Vedaldi and Zisserman for classification purposes [25]. As soon as the classification is finished, an evaluation of the category is computed using the mean Average Precision (AP) unit which will output four figures: a result containing the top 36 images with the best scores from the selected category and a graph displaying the AP for the training stage, and two for the testing stage. The extensive testing is performed on state-of-the-art feature extractors and datasets which will provide the full understanding of the system's behavior.

### 2.1.1 Convolutional Neural Networks

CNN models stand for one of the oldest deep neural networks hierarchy that have hidden layers among the general layers to help the weights of the system learn more about the features found in the input image. A general CNN architecture looks like the one shown in Figure 2 and consist of distinct types of layers. The convolutional layer applies an array of weights to all of the input sections from the image and creates the output feature map. The pooling layers simplify the information that is found in the output from the convolutional layer. The last layer is the fully connected layer that oversees the gathering of the findings from former layers and provides an N-dimensional vector, where N stands for the total number of classes. In the context of materials, it gives information on the texture of the material. Figure 3 gives a more descriptive identification of all the components of the pipeline, in particular the CNN architectures that this project uses for the challenging task of material classification. Both standard and state-of-the-art models have been used in the Caffe framework [36] in MATLAB. This subsection focuses on delivering a detailed explanation of the structure of each CNN.

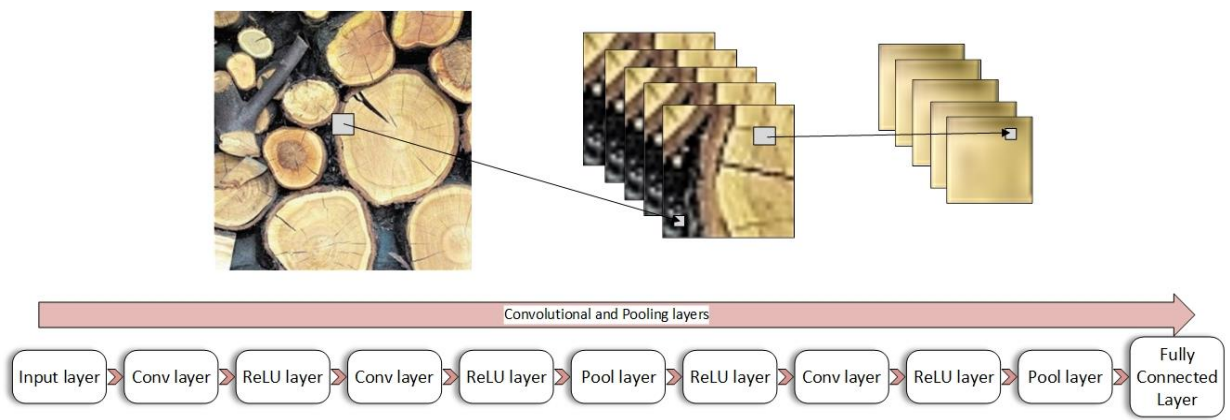

Figure 2 – A General CNN layer hierarchy

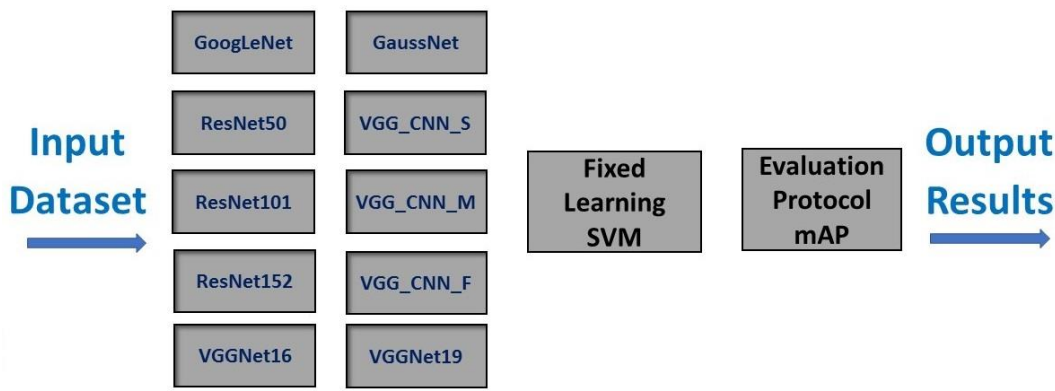

Figure 3 - Pipeline including all selected CNN architectures and SVM

The first CNN architecture is the VGG_CNN_F model [3] that has eight learned layers with five convolutional and two fully-connected, where "F" stands for fast. It is called the fast model due to the first layer having a stride of four pixels with 64x11x11 dimensions. The last layer gives an output encompassing around 4096 features detected in the input image that are further passed along to the classifier. The second CNN architecture is the VGG_CNN_M model [4] that it is called medium because its algorithm trains slower. This is due to the first layer having a stride two pixel that will change and decrease down to a stride one pixel in its third convolutional layer. Same as the CNN_F architecture, the last layer also outputs 4096 features. The third CNN is the VGG_CNN_S [37] that has a similarly built structure like the abovementioned architectures, which the distinction of the stride decreasing to the second convolutional layer instead of the third convolutional layer. This affects the training phase by slowing it down. These architectures can be studied in Figure 4.

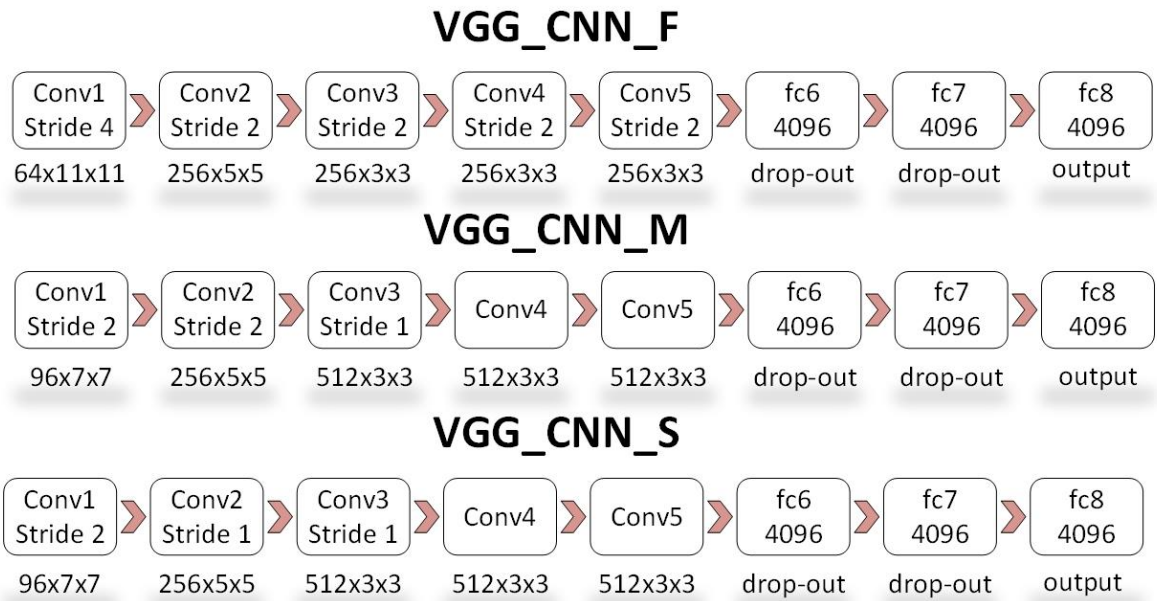

Figure 4 – VGG_CNN architectures overview

With the research focusing on CNN architectures, shortly after the abovementioned CNN models were published, a new state-of-the-art architecture called GoogLeNet [18] entered the scene and provided an improved network with larger depths and widths as well as fewer parameters used. With its 22 layers, the lower layers are formed of convolutional layers used to solve the performance issue caused by the expansion in the depth of the network, and the higher layers consist of the pooling layers which are introduced at all stages in the network and 9 brand-new type of hierarchies called inception that are modules stacked one upon another used for dimension reduction and can be up to 100 layers each. Compared to VGG_CNN_F [3] it uses less parameters within its network and offers an alternative to using fully connected layers. Since it presents a new type of architecture where models do not need to have the layers structured sequentially, it is extremely important that it is one of the CNN models tested. The fifth and sixth CNN models used are VGG16 and VGG19 [34] very similar to the first three types of architectures consisting of eight learned layers. The new models test their deeper network on the ImageNet7 dataset and achieve better results than GoogLeNet and VGG_CNNs where the architecture's test error is around 7.0% compared to 7.9% outputted by GoogLeNet. Two noticeable characteristics are: i) the stride is fixed to one pixel across the majority of the network and ii) the convolutional layer is not followed by a pooling layer at all stages in the hierarchy. Although the configuration of the fully connected layers is the same in all networks, the first two levels gather 4096 features, while the third one only has 1000 channels. Since these models are mainly trained and tested on only one database, the question is whether they will achieve similar results on other datasets and generalize as good for other tasks. Furthermore, a data augmentation technique has been used during the training of the network. The structure of these three architectures can be seen in Figure 5.

The state-of-the-art ResNet50, ResNet101 and ResNet152 architectures introduce more complex types of networks as deeps as 152 layers that reaches a much lower error rate of 3.57% on the ImageNet7 dataset [35]. Despite obtaining very superior results, they also bring up a new issue: degradation, and they question whether the learning process gets better with the addition of multiple layers. It presents a comparable structure to VGG, but has supplementary shortcut connections. The model has five convolutional layers and ends with a pooling layer with 1000 channels. The CNNs structure can be studied in Figure 6.

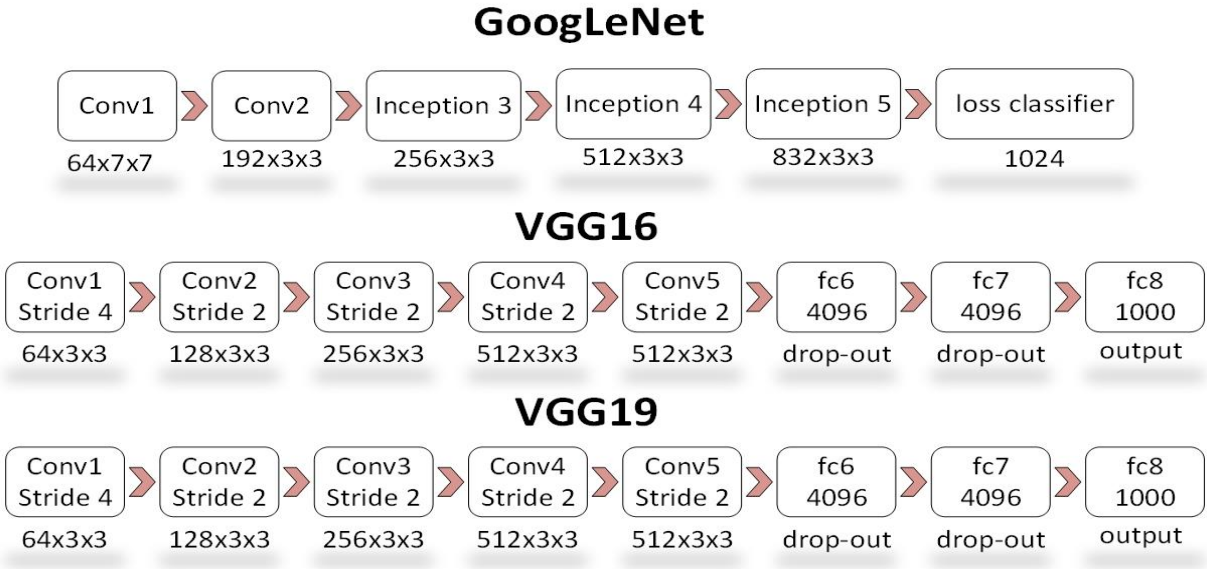

Figure 5 – GoogLeNet and VGG architectures overview

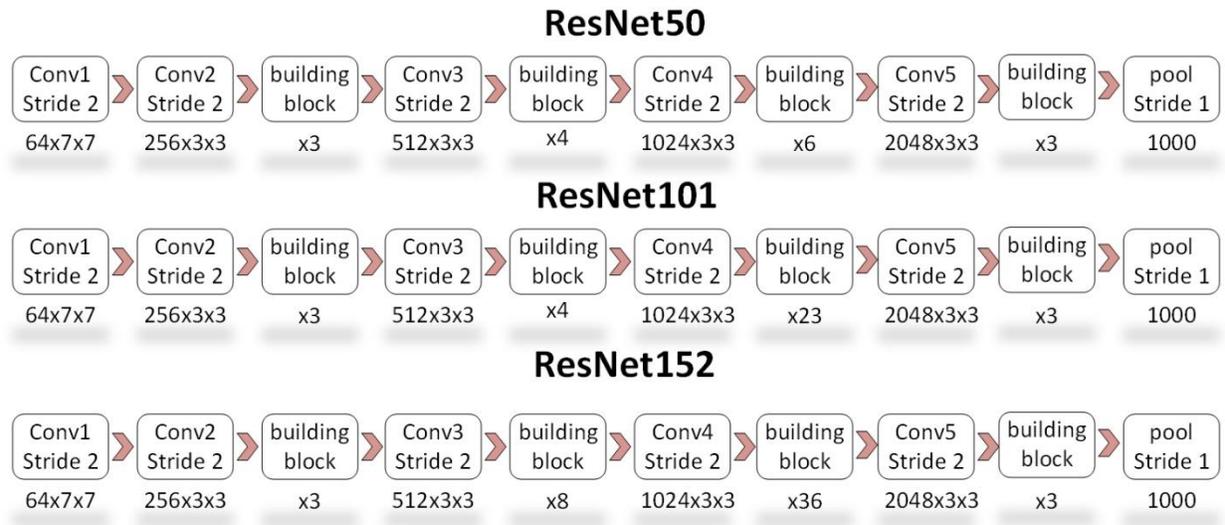

Figure 6 – ResNet architecture overview

The differences in the structure of all the architectures used for the experimental tests lead to the avoidance of bias being created as well as making sure that all the combinations between the CNN models and datasets are varied and diverse. The results extracted from the last layer are further given to the next part of the system, the classifier.

2.1.2 Support Vector Machine

The classifier used for this task is a linear Support Vector Machine (SVM) [25]. An SVM is used mainly in supervised learning in tasks such as image classification and regression problems. Being memory efficient and working best with challenging classes, it can identify the closest features or points in the images and produce support vectors containing those points. The next step for the algorithm after it finds these points is to draw a line between the closes positions and conclude that the line that bisects the initial drown link is where the two classes are found [26]. The original drawn line is what finds the distance between the two established classes. This similarity functionality is known as a linear kernel that achieves as good results as other non-linear or multilinear kernels [27]. Important results have been brought linear SVMs into focus when Tang [28] proved that linear classifiers work better than other functions employed on handwritten digit classification and face recognition which offered the beginning of extended research. After the CNN model extracts all the features distinguished in the image, a label of +1 or -1 is assigned to each picture that will indicate whether it belongs to positive or negative training or testing. These labels are put in the abovementioned support vector and are calculated through the formula:

$$K(h, h') = \sum_{k=1}^{n} h_k h'_k$$

to calculate the scores for each individual image. According to the scores received based on the weights in the vector, the images are ranked and the algorithm will choose the top 36 best-ranked pictures to be displayed to the user. Moreover, the ranked images are passed over to the evaluation protocol that uses the results to perform the comparison and outline the performance of the feature extractor on the dataset assigned as training and testing. A linear SVM has been implemented due to popular reasons like not overfitting the data too much, when the number of features is large and it achieves overall small timings for the computation of the tests. Since four distinct datasets with several images per category ranging from 100 to 2500 ware chosen, this was the better choice that fit the situation of material classification.

## 2.2 Approaches

Training a deep network to identify distinct types of materials has received a lot of attention in recent years since the issue of localizing and recognizing an object can be challenging when applied to real-world images. This section outlines the approaches that this project applied to the overall system and emphasizes the changes that appear due to each method. To make sure that a thorough assessment is done at all times during the extensive testing and the tendency of having a bias is not formed, every component of the system has been examined and selected to a high standard where the combinations between the elements of the pipeline are diverse and provide good evidence towards the task. The results from these approaches are presented in Technical Achievement section.

### 2.2.1 Datasets

Four real-world datasets have been selected to conduct the experimental tests. In order to adapt the datasets to the algorithms in MATLAB, a short period of time was spent at the beginning of the project to make sure that every image has the correct format of JPEG as well as the right name. These changes lead to a smooth running of the tests. The first dataset is the Flickr Material Database (FMD) [10] that captures the real-world appearance of materials and was developed with the initial intention to study human material categorization, but has become the benchmark for material recognition instead. It uses the Flickr API to collect 100 images, 50 close-ups and 50 regular views, per category for 10 categories. The images were manually gathered under conditions that ensured to have information such as illumination, color, texture or compositions. The ten categories can be noticed in Figure 7.

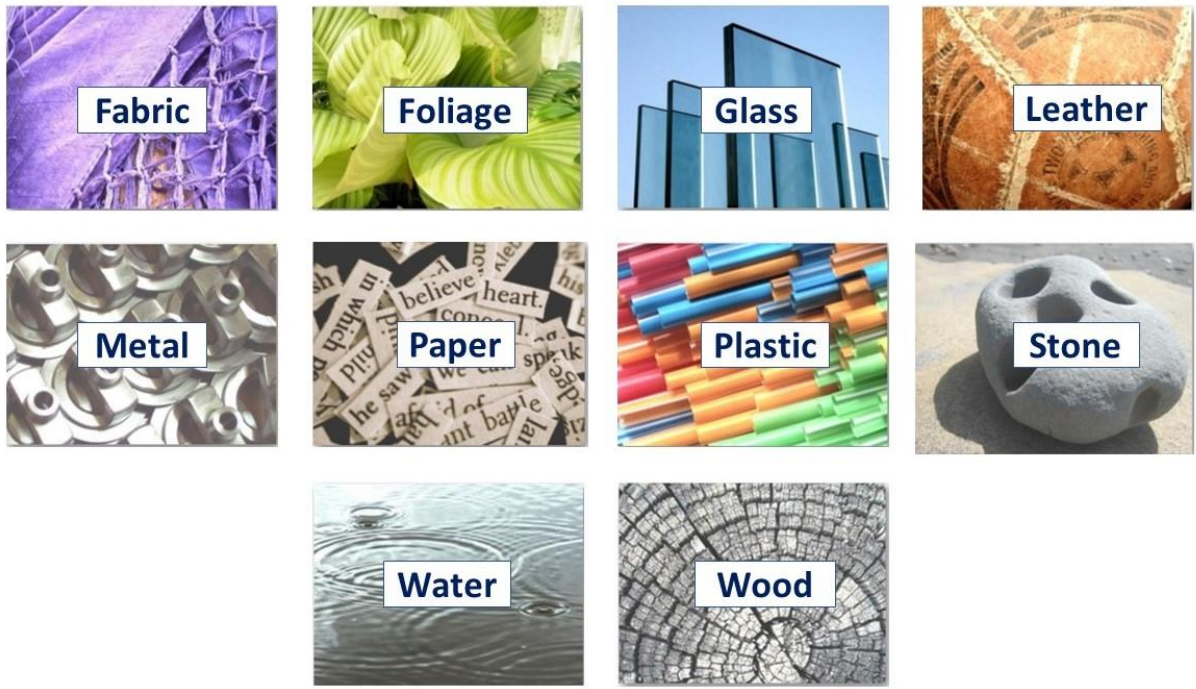

Figure 7 – FMD categories

The second dataset is the ImageNet-Material7 (ImageNet7) published by Deng et al [29] that is structured after the WordNet hierarchy and it is a subset of ImageNet2010 database introduced in [30]. The original motivation was providing researchers with a very large dataset where each photograph is manually annotated and quality-controlled. It consists of seven categories that can be observed in Figure 8, each category having 1000 images. The dataset offers 7000 images in total obtained from a variety of sources that is the base that Russakovky et al. [31] used to train the algorithms as well as using their protocols to further annotate new images. FMD and ImageNet7 have become two standard databases that are very similar in the way the materials are captured in the images. Schwartz and Nishino [32] compare the two datasets on the task of per-pixel recognition of materials and conclude that ImageNet7 outperforms FMD achieving an accuracy of 60.5%.

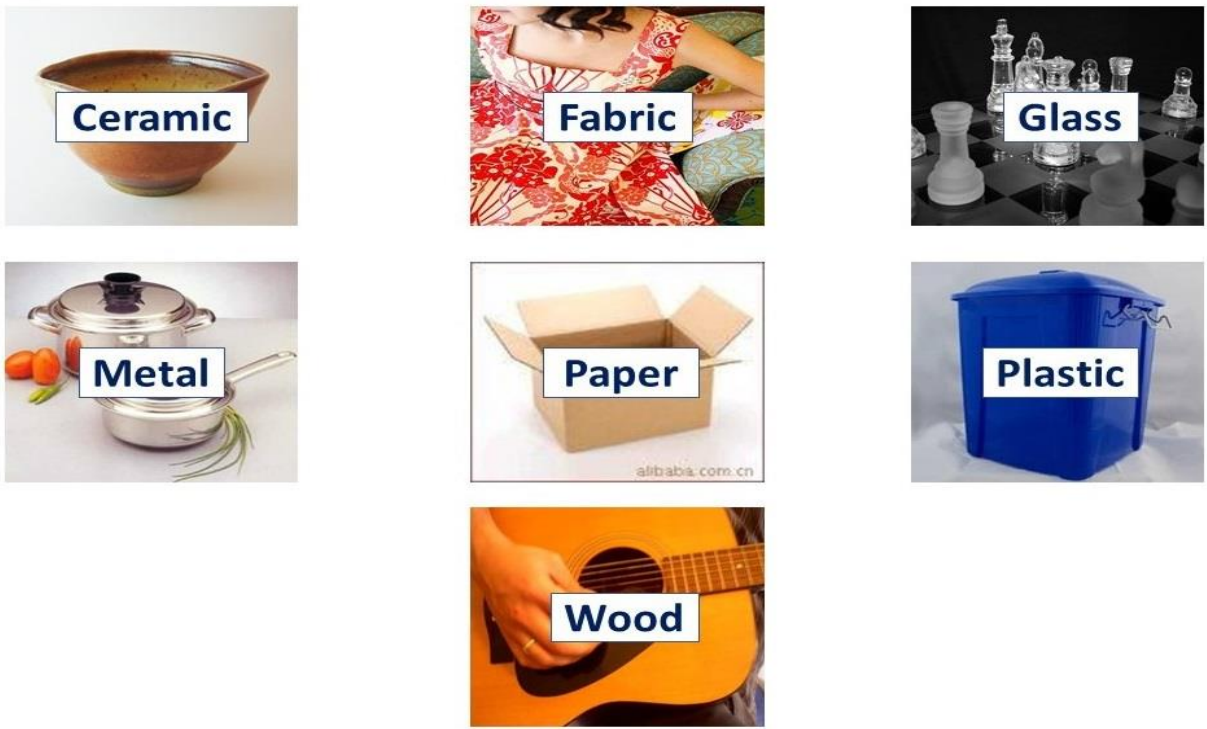

Figure 8 – ImageNet7 categories

The third dataset is the Materials in the Context Database (MINC2500) [19] that presents a larger dataset with a total of 23 well-sampled categories for patch material classification and material recognition. MINC2500 is a subset of the MINC dataset and has 2500 images per category, offering a very diverse and complex database for materials extracted from the real-world. When it was published, the dataset outperformed all earlier attempts by using state-of-the-art convolutional neural network architectures and obtaining an accuracy of 85.2%. The dataset can be observed in Figure 9.

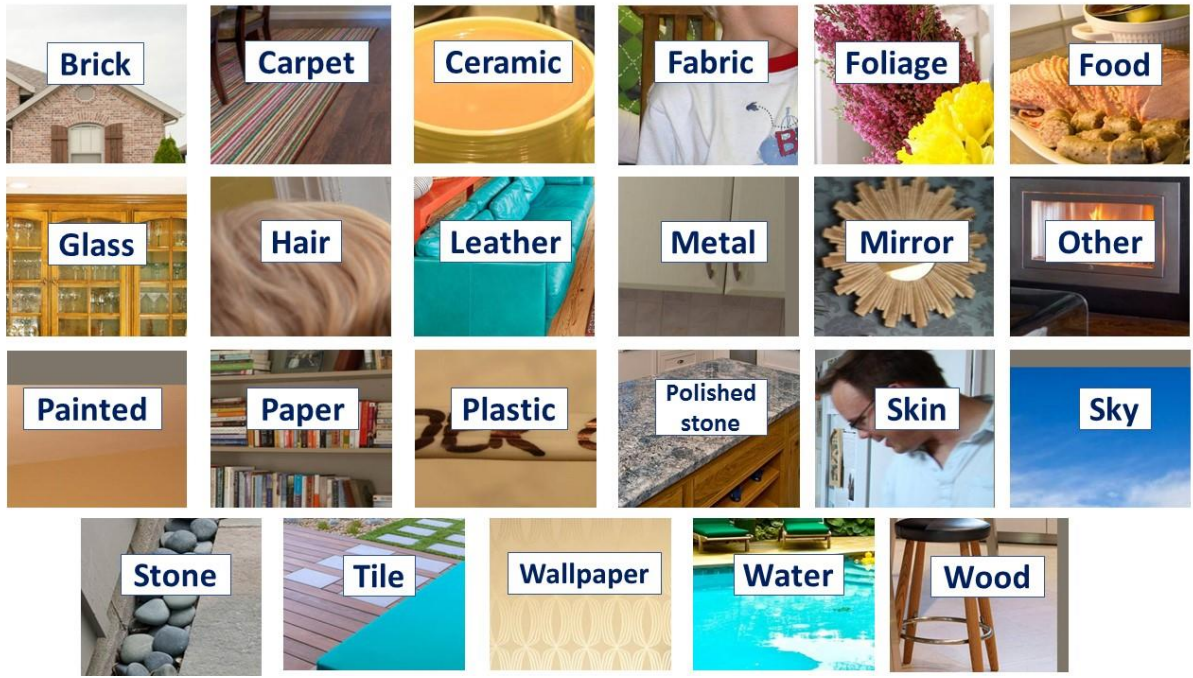

Figure 9 – MINC2500 categories

The last dataset is the Google Material Dataset (GMD) [23] that was published recently and it is the newest for material classification. It has around 10k manually cropped and annotated images divided into 10 categories. The different thing about this dataset is that the images are captured in real-world scenarios and therefore they will differ in illumination and view compared to other datasets, but are built on the same principle of close-up and general view. [23] gives a comparison between FMD and GMD when the datasets are divided into training, validation and testing phases and where the newly introduced material dataset achieves an accuracy of 74% compared to FMD that achieves 64%. One noticeable fact about GMD is that the number of images in each category are different which leads to some classes performing better and others performing worse due to the inexact number of images overall. This results in the mean Average Precision unit for the feature extractor not being able to be computed on the results, since every category holds a distinct number of pictures. The number of images per categories ranges from 324 images for paper up to 2134 for fabric which already outlines that fabric will have a much better outcome. This database can be seen in Figure 10 and an overview across all datasets is emphasized in Figure 11.

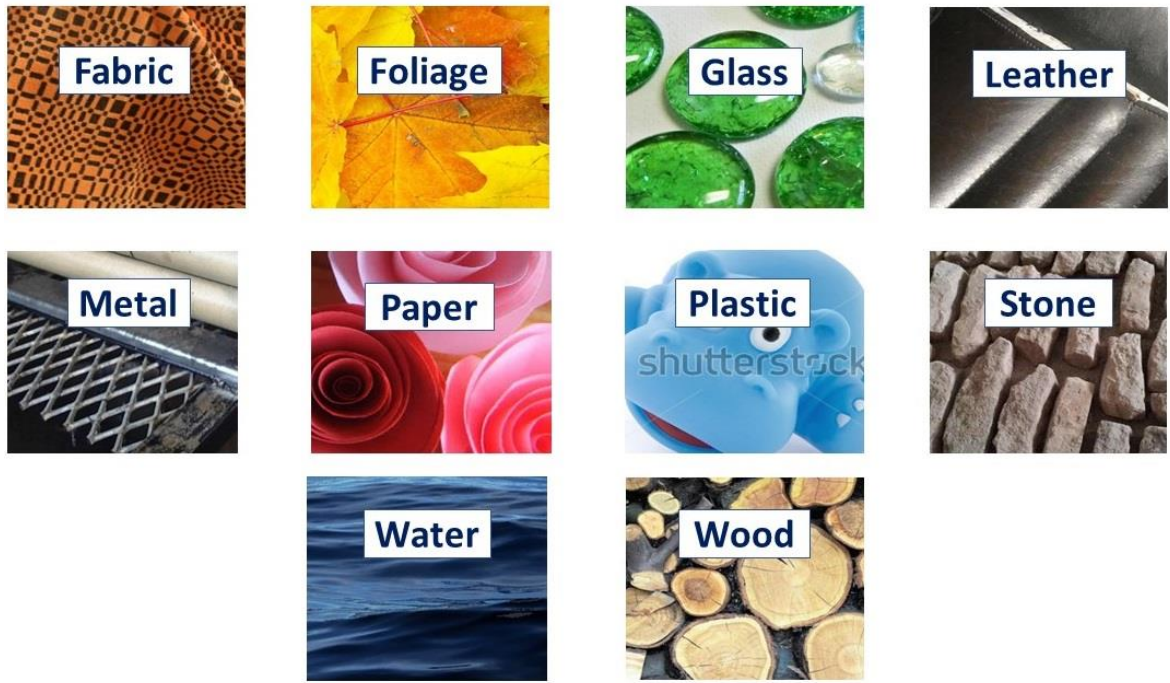

Figure 10 – GMD categories

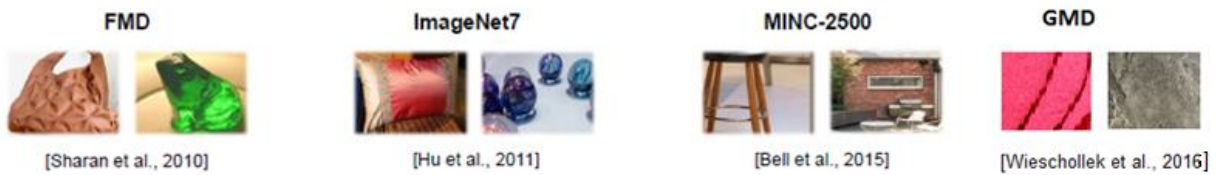

Figure 11 – a) Table consisting of datasets used in the system's pipeline; b) figures of the chronological order of the publishing of the datasets

As it shown in Figure 11 and has been outlined above, each dataset has a different number of images per category. This diversity is meant to avoid bias from occurring during the tests. For each category across all datasets, the first half of the images are used for the positive training and the remaining half of the images are used for the positive testing. The negative testing set [25] of images is constant across all databases having images of animals that has been adapted to have the correct number of pictures for each dataset. For instance, when tests are run on FMD, it only has 50 images, while when it is run on MINC it has 1250 images. The negative training set is formed in the following way: for ImageNet7 database for the fabric category, it will contain 10% of the remaining classes (foliage, glass, leather, metal, paper, plastic, stone, water and wood). The reason behind this is, that in this way, the category still holds images from the same dataset, but different from the class that is being evaluated during the test.

### 2.2.2 Transfer learning

Transfer learning in the context of this project's pipeline implies inter-layer transfer of information and the fact that the information extraction happens at the layer before the last fully connected layer. This has been preferred in order to understand and analyze whether the influence of reflectance and shading information affects the accuracy of the feature extractor and outline if these two factors can be considered two main characteristics that help to decide the category the material belongs to. Wieschollek et al [23] introduces transfer learning and questions how it affects the system when FMD and GMD dataset are selected as inputs. The success of 1.2 million labelled images and the considerable number of distinct classes in the ImageNet dataset [30] have led researchers to investigate how transfer learning will affect earlier results obtained on the ImageNet7 subset [29, 33]. From their experimental work, using the AlexNet [3] CNN model they offered outcomes like pre-training with fewer classes and more data per category producing better results or naming a couple of the factors that influence transfer learning. Inspired by their work, this project aims at using transfer learning with deeper Convolutional Neural Networks architectures like VGG [34], ResNet [35] or GoogLeNet [18] and others to verify whether their findings are consistent and can be replicated using other models. Another interesting question to be answered proposed in [23] is whether the fact that the more training data a computer vision system receives and is trained on, either the better the results are, or it may provide the feature extractors with too many similar good features. The results obtained on the current pipeline have not been fine-tuned yet.

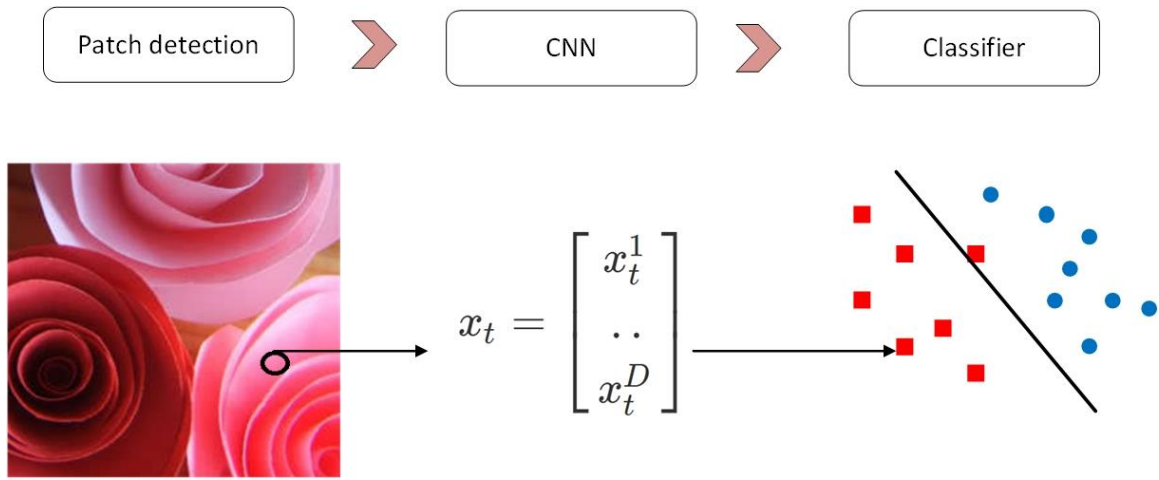

Figure 12 – Patch segmentation process

### 2.2.3 Patch segmentation

Patch segmentation stands for a technique that identifies what is at the center of a given image patch. Applied to material classification, it refers to recognizing the material from the localization of the central pixel in an image segment and categorizing it into the set of categories. The idea was used in the context of material classification in [19] where two important approaches: patch segmentation and full scene material classification have been emphasized and tested for three classifiers: AlexNet [3], VGG16[34] and GoogLeNet [18]. Patches are labelled segments that can be used to train a system. The way it is used is that the image center and patch scale are combined to define the entire patch region that is fed to the feature extractor algorithm. The MINC2500 dataset that this approach was initially tested on, contains images ready for patch classification and an interesting direction would be to observe whether state-of-the-art models, as well as other architectures, will give better or worse results.

# Chapter 3 Technical Documentation and Results

To increase the computational speed, the code has been changed to use all the GPUs installed in the computer allowing for a faster running of the tests. This section outlines the analysis of the results and achievements produced throughout the project. At the beginning of my project, I helped with the work for the [38] paper where the VGG_CNN_F, VGG_CNN_M and VGG_CNN_S are evaluated on three material datasets using data augmentation for the training of the learning system. There are two kinds of images used: real-world images and greyscale versions of the real-world images to which three types of augmentations are applied: no augmentation, crop and flip augmentation and no collation. The results show that if data augmentation is used, the outcomes can be improved achieving up to 94.99% for the challenging task of classifying materials. As the Average Precision was only able to be computed only on three of the four selected databases, the results will be shown in different tables and images.

## 3.1 Evaluation on common ground

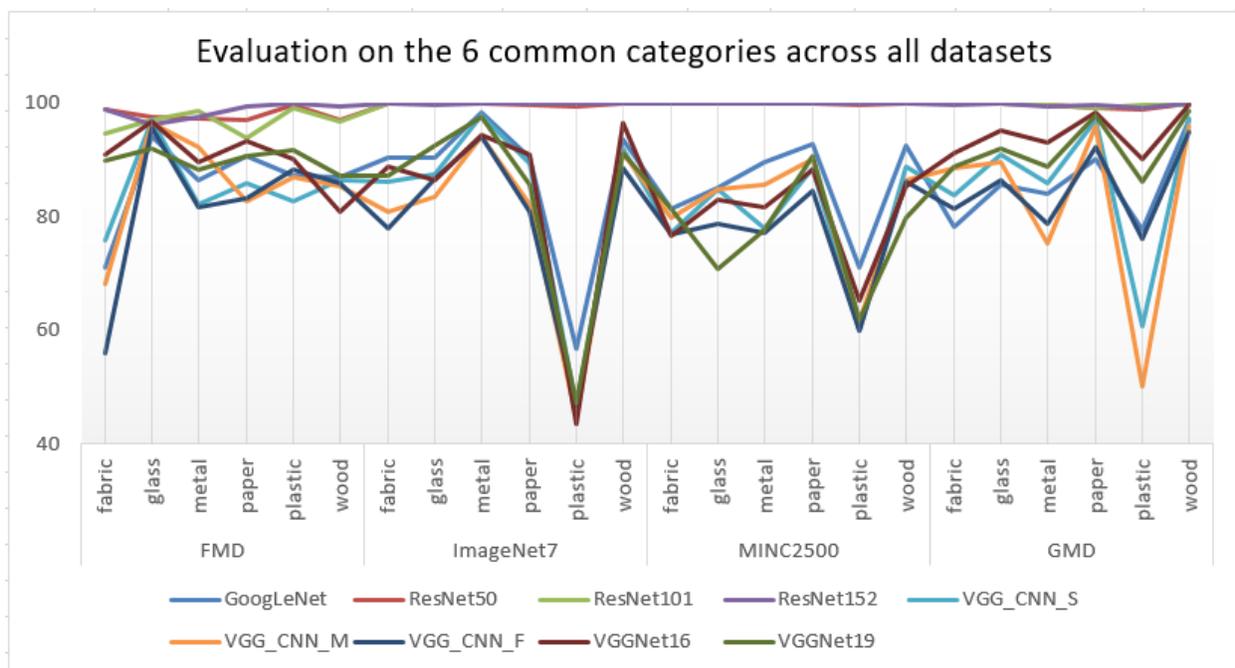

Figure 13 – Architecture evaluation

Figure 13 denotes the evaluation of the experimental tests that have been done on nine CNNs on four chosen datasets: FMD, ImageNet7, MINC and GMD. The above graph is created on the six common categories across all datasets: fabric, glass, metal, paper, plastic and wood. Looking at the common ground on which these feature extractors are trained and tested on will determine the way in which each of these databases will provide a distinct perspective on the overall system and how it affected by such factors. As it is shown in the graph in Figure 13, the best-recognized categories are paper and metal by the majority of the CNNs, and the worst recognized category is plastic across all datasets. The current system succeeds in a good general recognition of the common categories which is emphasized by looking at the accuracies in Table 1 where both paper and metal classes obtain an average accuracy of ~91.2-92.7 across the trained feature extractor. As some categories are more difficult to distinguish than others due to the interest points not being labelled correctly or the lack of training data and a targeted classifier, the system only manages to achieve ~78.5 for the plastic category. One important observation is that the constant results achieved by the ResNet CNNs across all categories and datasets lead to only one explanation available for this situation: the implementation of the ResNet 50, 101 and 152 layers with 7x7 convolutional kernels to a 3x3 convolutional kernels is too small and therefore, it affects the way the features are extracted from the input images. This fact has been concluded since the architecture's results are constant within the same range across all tested categories and they do not vary like the other CNNs in their adaptation to detecting the features.

For the remaining of this section, the ResNet architectures will not be considered for comparison purposes since they are not producing precise findings in the context of material classification. On the FMD dataset, VGGNet16 produces the best mean Average Precision at ~90.2, result which is supported by the network's structure of 3x3 convolutional kernels with a stride of 1 fixed across the entire network that leads to little to no information being lost. The next dataset trained and tested offers more data which allows for a deeper network architecture oriented such as GoogLeNet to outperform VGGNet16 on this database and obtain an accuracy of ~86.5 compared to 83.3%. By adding more data to the training phase of the learning process, the same configuration as on ImageNet uses its inception modules to improve its feature detection algorithm and outcomes ~85.3 mAP. All of the abovementioned results are shown in Table 2. The GMD has different amounts of images for each category: fabric – 1067, foliage - 531, glass – 315, leather – 630, metal – 439, paper – 162, plastic – 183, stone – 681, water – 231 and wood – 605 images.

Table 1 – Evaluation results on the common ground categories

| Dataset | FMD | | | | | | ImageNet7 | | | | | |
|---|---|---|---|---|---|---|---|---|---|---|---|---|
| Category | fabric | glass | metal | paper | plastic | wood | fabric | glass | metal | paper | plastic | wood |
| GoogLeNet | 71.07 | 94.03 | 86.3 | 90.57 | 87.25 | 86.87 | 90.41 | 90.4 | 98.28 | 90.29 | 56.67 | 93.47 |
| ResNet50 | 98.9 | 97.59 | 97.31 | 97.15 | 99.65 | 96.98 | 100 | 99.8 | 100 | 99.68 | 99.52 | 100 |
| ResNet101 | 94.56 | 97.15 | 98.52 | 93.88 | 99.04 | 96.8 | 100 | 99.8 | 99.82 | 99.98 | 99.83 | 100 |
| ResNet152 | 98.83 | 96.36 | 97.59 | 99.5 | 99.83 | 99.34 | 100 | 99.6 | 100 | 99.97 | 99.97 | 100 |
| VGG_CNN_S | 75.7 | 97.07 | 82.19 | 85.8 | 82.61 | 86.49 | 86.08 | 87.5 | 97.78 | 89.34 | 46.64 | 92 |
| VGG_CNN_M | 68 | 96.54 | 92.28 | 82.69 | 87 | 85.35 | 80.78 | 83.5 | 94.17 | 82.21 | 44.47 | 91.22 |
| VGG_CNN_F | 55.96 | 95.75 | 81.61 | 83.3 | 88.22 | 85.92 | 78.01 | 86.6 | 94.07 | 80.74 | 46.68 | 88.52 |
| VGGNet16 | 90.92 | 96.68 | 89.66 | 93.21 | 90.14 | 80.79 | 88.69 | 86.5 | 94.28 | 90.86 | 43.35 | 96.48 |
| VGGNet19 | 89.79 | 91.95 | 88.38 | 90.68 | 91.75 | 87.15 | 87.14 | 92.5 | 97.65 | 85.66 | 47.09 | 91.15 |
| Dataset | MINC2500 | | | | | | GMD | | | | | |
| Category | fabric | glass | metal | paper | plastic | wood | fabric | glass | metal | paper | plastic | wood |
| GoogLeNet | 81.39 | 84.97 | 89.51 | 92.87 | 71.04 | 92.52 | 78.17 | 85.5 | 84.04 | 90.23 | 77.61 | 96.91 |
| ResNet50 | 99.97 | 99.95 | 99.98 | 99.99 | 99.81 | 100 | 99.99 | 100 | 99.65 | 99.21 | 98.82 | 99.95 |
| ResNet101 | 99.99 | 99.96 | 99.87 | 99.92 | 99.91 | 99.95 | 99.92 | 99.9 | 99.67 | 99.07 | 99.57 | 99.75 |
| ResNet152 | 100 | 99.9 | 99.98 | 99.96 | 99.99 | 99.95 | 99.71 | 99.8 | 99.41 | 99.73 | 99.23 | 99.91 |
| VGG_CNN_S | 77.21 | 84.76 | 78.07 | 88.66 | 59.81 | 88.84 | 83.87 | 90.9 | 85.89 | 97.07 | 60.76 | 97.19 |
| VGG_CNN_M | 79.85 | 84.76 | 85.51 | 90.15 | 61.39 | 86.47 | 88.52 | 89.5 | 75.34 | 95.86 | 50.01 | 95.95 |
| VGG_CNN_F | 76.85 | 78.84 | 77.16 | 84.67 | 59.97 | 86.05 | 81.37 | 86.4 | 78.72 | 92.29 | 76.18 | 95.04 |
| VGGNet16 | 76.73 | 83.03 | 81.7 | 88.27 | 65.32 | 85.48 | 91.15 | 95.2 | 93.06 | 98.43 | 90.14 | 99.75 |
| VGGNet19 | 81.36 | 70.68 | 77.73 | 90.8 | 61.76 | 79.88 | 88.76 | 92.1 | 88.79 | 97.94 | 86.05 | 98.55 |

Table 2 – mAP achieved on the common categories

| Dataset | FMD | ImageNet7 | MINC2500 |
|---|---|---|---|
| GoogLeNet | 86 | 86.5 | 85.3 |
| ResNet50 | 97.9 | 99.8 | 99.9 |
| ResNet101 | 96.6 | 99.9 | 99.9 |
| ResNet152 | 98.5 | 99.9 | 99.9 |
| VGG_CNN_S | 84.9 | 83.2 | 79.5 |
| VGG_CNN_M | 85.3 | 79.3 | 81.3 |
| VGG_CNN_F | 81.7 | 79.1 | 77.2 |
| VGGNet16 | 90.2 | 83.3 | 80 |
| VGGNet19 | 89.9 | 83.5 | 77 |

As it is observed only from the numbers of images per category, it can be outlined even before performing the tests that some categories will produce more accurate results due to a higher number of data available for the system. This limitation leads to the evaluation unit not being able to be used in this dataset's context, therefore it will be analyzed by the accuracy per category per CNN. In Table 1, the GMD section of the table emphasizes that the top three categories are in the following order: wood, paper and glass. The findings are very interesting as if one is to look at the number of images per category, most of the top found classes do not hold the highest number of data and these results emphasize that one does not always need more images if the feature extractor is able to extract the feature vector correctly so that the classifier can compute the score map and compute the conclusions. Evaluating on only the common categories allows an exact comparison between all the components of the system without having the influence of other factors.

## 3.2 Evaluation of all categories

This subsection looks at all the results gathered across all datasets and outlines the research contribution presented through the experimental results. Although some of the above results are incorporated in the outcomes discussed in this section, the analysis presented is done generally and includes all categories and experimental tests performed. To take advantage of all of the data available, all categories from all datasets have been used for training and testing and have been included in the evaluation. Therefore, all 23 categories of MINC-2500, all 7 categories of ImageNet, all 10 categories of FMD and all 10 categories of GMD have been used as inputs for the system created.

By looking at Table 3, it is observed that the best performing CNN architecture is on MINC-2500 dataset at ~92.56, followed by the VGGNet16 model which manages to achieve an accuracy of ~91.86 on FMD dataset and followed by VGGNet19 on ImageNet7 with approximately 79.08%. Being the largest real-world dataset having the most images out of the four selected databases, it adapts best at the task of classifying materials and succeeds in achieving the best performance overall when state-of-the-art models are used. When comparing the existing results on the MINC2500 dataset with the ones obtained by Bell et al in [19], there are some important observations to be made. Firstly, in the published paper, only three types of models were tested: AlexNet, GoogLeNet and VGG16, whereas my project introduces a total of nine variety of models including GoogLeNet and VGG16 and attempts at observing how each set of experiments will compute better features.

Table 3 – Evaluation results for all categories on three datasets

| CNN | MINC-2500 | ImageNet7 | FMD |
|---|---|---|---|
|  | mAP | mAP | mAP |
| GoogLeNet | **92.56** | 58.78 | 88.32 |
| ResNet50 | 99.92 | 99.79 | 97.65 |
| ResNet101 | 99.93 | 99.85 | 97.04 |
| ResNet152 | 99.94 | 99.83 | 98.16 |
| VGG_CNN_S | 89.09 | 78.55 | 86.86 |
| VGG_CNN_M | 89.91 | 75.48 | 86.8 |
| VGG_CNN_F | 88.14 | 74.97 | 85.58 |
| VGGNet16 | 89.14 | 78.97 | **91.86** |
| VGGNet19 | 87.76 | **79.08** | 91.56 |

The results obtained suggest that GoogLeNet combined with the current pipeline and transfer learning manages to improve on their results of 85.2% with a ~7.3% increase. When looking at the same dataset, but in the context when VGGNet16 is used, the current network outperforms the previous results as well by achieving ~89.1 accuracy which represents a 4.3% improvement. These results outline the fact that the popular technique of using transfer learning affects the overall training of the network that leads to a better classification of the interest points and their correlation to each class for obtaining a better score map.

FMD achieves the best performance with a VGGNet16 of around 91.86 accuracy that is a deep network with 16 layers. Compared to [34], where it was outlined that this type of model achieved very superior results on the ImageNet dataset, these outcomes answer the question whether the preliminary conduct of the model can be replicated on other datasets. If one is to look at earlier work done by Hu et al. [29] where, by using kernel descriptors, a 54% accuracy on the dataset was obtained, or, more recently at the work done by Kalliatakis et al. [38] that improved the accuracy by 10.4% by using the VGG_CNN-M with data augmentation, the networks seem to become better at understanding the way the information should be extracted from the convolutional kernels to be transmitted to the classifiers to correctly identify the labels. By using a deeper network where each layer transfers information from the other neighboring layers to achieve a better learning of material characteristics and the loss of information being to a minimum due to the stride being 1 at the majority of the layers in the network, the existing configuration tends to compute better results across all categories tried. Furthermore, one potential explanation can be that the small size of the datasets may show the presence of overfitting in this context.

The remaining dataset is ImageNet7 which obtains 79.08% accuracy on the deep model VGGNet19. Although the performance is not as good as the other sources of images since it has fewer number of images per category, compared to 60% achieved in [29] and 74.97% achieved in [38], it outlines that deep networks are much better at classification tasks especially if transfer learning has been used as an approach. In this type of architecture can be suggested the fact that by adding multiple layers the performance can be improved. As it is observed in Table 3, the deeper and more layers a model has, the more accurately the feature vector will be constructed and the better the classifier will recognize the type of material from the respective categories. In Table 4, Table 5 and Table 6, it can be observed the accuracy of all the nine CNNs run on MINC2500, ImageNet7 and FMD databases. Average Precision has been used on all the results from these three tables and produced the results in Table 3.

Table 4 – MINC2500 dataset results

| Dataset | MINC2500 | | | | | | | | | | |
|---|---|---|---|---|---|---|---|---|---|---|---|
| Category | brick | carpet | ceramic | fabric | foliage | food | glass | hair | leather | metal | mirror |
| GoogLeNet | 96.96 | 94.42 | 89.49 | 81.39 | 95.4 | 98.73 | 84.97 | 97.21 | 95.84 | 89.51 | 94.28 |
| ResNet50 | 100 | 99.99 | 99.79 | 99.97 | 99.98 | 99.85 | 99.95 | 99.95 | 99.98 | 99.98 | 99.96 |
| ResNet101 | 99.96 | 99.93 | 99.95 | 99.99 | 100 | 99.97 | 99.96 | 99.97 | 99.98 | 99.87 | 99.92 |
| ResNet152 | 99.93 | 99.89 | 99.86 | 100 | 99.95 | 99.89 | 99.9 | 99.93 | 99.95 | 99.98 | 99.97 |
| VGG_CNN_S | 95.95 | 87.57 | 87.37 | 77.21 | 88.81 | 97.53 | 84.76 | 93.86 | 94.01 | 78.07 | 93.88 |
| VGG_CNN_M | 96.66 | 91.38 | 88.7 | 79.85 | 90.05 | 97.37 | 84.76 | 92.25 | 94.61 | 85.51 | 93.13 |
| VGG_CNN_F | 95.44 | 88.35 | 85.75 | 76.85 | 87.67 | 96.98 | 78.84 | 92.04 | 95.35 | 77.16 | 94.01 |
| VGGNet16 | 93.39 | 83.69 | 91.12 | 76.73 | 80.78 | 97.32 | 83.03 | 95.5 | 94.71 | 81.7 | 93.06 |
| VGGNet19 | 95.34 | 79.34 | 89.89 | 81.36 | 87.73 | 97.56 | 70.68 | 96.53 | 93.96 | 77.73 | 94.13 |

| Dataset    | MINC2500 | | | | | | | | | | | |
|---|---|---|---|---|---|---|---|---|---|---|---|---|
| Category   | other | painted | paper | plastic | polished stone | skin | sky | stone | tile | wallpaper | water | wood |
| GoogLeNet  | 89.44 | 94.1 | 92.87 | 71.04 | 97.05 | 96.9 | 99.33 | 92.89 | 93.08 | 96.7 | 94.81 | 92.52 |
| ResNet50   | 99.99 | 99.93 | 99.99 | 99.81 | 99.87 | 99.81 | 99.96 | 99.97 | 99.96 | 99.97 | 99.46 | 100 |
| ResNet101  | 99.86 | 99.88 | 99.92 | 99.91 | 99.98 | 99.96 | 99.86 | 99.99 | 99.97 | 99.98 | 99.71 | 99.95 |
| ResNet152  | 99.98 | 99.97 | 99.96 | 99.99 | 99.98 | 99.96 | 99.95 | 99.97 | 99.98 | 99.86 | 99.83 | 99.95 |
| VGG_CNN_S  | 82.49 | 90.93 | 88.66 | 59.81 | 94.65 | 93.73 | 99.16 | 90.14 | 92.11 | 95.5 | 94.14 | 88.84 |
| VGG_CNN_M  | 82.59 | 92.6 | 90.15 | 61.39 | 93.63 | 94.12 | 98.82 | 92.43 | 93.58 | 94.37 | 93.55 | 86.47 |
| VGG_CNN_F  | 85.1 | 92.99 | 84.67 | 59.97 | 94.2 | 90.64 | 98.91 | 90.45 | 87.89 | 93.62 | 94.38 | 86.05 |
| VGGNet16   | 85.02 | 92.61 | 88.27 | 65.32 | 94.37 | 95.45 | 98.81 | 91.39 | 94.05 | 96.79 | 91.68 | 85.48 |
| VGGNet19   | 78.88 | 86.67 | 90.8 | 61.76 | 95.34 | 91.95 | 98.96 | 87.87 | 94.02 | 95.58 | 92.44 | 79.88 |

Table 5 – ImageNet7 dataset results

| Dataset    | ImageNet7 | | | | | | |
|---|---|---|---|---|---|---|---|
| Category   | ceramic | fabric | glass | metal | paper | plastic | wood |
| GoogLeNet  | 52.8 | 90.41 | 90.44 | 98.28 | 90.29 | 56.67 | 93.47 |
| ResNet50   | 99.5 | 100 | 99.82 | 100 | 99.68 | 99.52 | 100 |
| ResNet101  | 99.5 | 100 | 99.82 | 99.82 | 99.98 | 99.83 | 100 |
| ResNet152  | 99.29 | 100 | 99.62 | 100 | 99.97 | 99.97 | 100 |
| VGG_CNN_S  | 50.53 | 86.08 | 87.48 | 97.78 | 89.34 | 46.64 | 92 |
| VGG_CNN_M  | 52.08 | 80.78 | 83.45 | 94.17 | 82.21 | 44.47 | 91.22 |
| VGG_CNN_F  | 50.14 | 78.01 | 86.64 | 94.07 | 80.74 | 46.68 | 88.52 |
| VGGNet16   | 52.58 | 88.69 | 86.54 | 94.28 | 90.86 | 43.35 | 96.48 |
| VGGNet19   | 52.43 | 87.14 | 92.46 | 97.65 | 85.66 | 47.09 | 91.15 |

Table 6 – FMD dataset results

| Dataset | FMD | | | | | | | | | |
|---|---|---|---|---|---|---|---|---|---|---|
| Category | fabric | foliage | glass | leather | metal | paper | plastic | stone | water | wood |
| GoogLeNet | 71.07 | 98.79 | 94.03 | 92.59 | 86.3 | 90.57 | 87.25 | 86.94 | 88.78 | 86.87 |
| ResNet50 | 98.9 | 97.45 | 97.59 | 98.16 | 97.31 | 97.15 | 99.65 | 98.76 | 94.56 | 96.98 |
| ResNet101 | 94.56 | 98.97 | 97.15 | 99.1 | 98.52 | 93.88 | 99.04 | 94.56 | 97.84 | 96.8 |
| ResNet152 | 98.83 | 98.68 | 96.36 | 96.8 | 97.59 | 99.5 | 99.83 | 99.55 | 95.07 | 99.34 |
| VGG_CNN_S | 75.7 | 94.05 | 97.07 | 95.96 | 82.19 | 85.8 | 82.61 | 82.81 | 85.91 | 86.49 |
| VGG_CNN_M | 68 | 92.62 | 96.54 | 87.88 | 92.28 | 82.69 | 87 | 85.19 | 90.44 | 85.35 |
| VGG_CNN_F | 55.96 | 96.31 | 95.75 | 93.11 | 81.61 | 83.3 | 88.22 | 86.77 | 88.88 | 85.92 |
| VGGNet16 | 90.92 | 96.79 | 96.68 | 94.47 | 89.66 | 93.21 | 90.14 | 91.45 | 94.48 | 80.79 |
| VGGNet19 | 89.79 | 95.85 | 91.95 | 96.56 | 88.38 | 90.68 | 91.75 | 90.54 | 92.99 | 87.15 |

GMD is the last dataset that experimental tests have run on. Due to the fact that each category has a different number of images, the AP unit was unable to be used which lead to the evaluation being a little bit different. Table 7 displays all the accuracies obtained for each of the ten categories of the dataset. The top three architectures based on the classifier scores are VGGNet16 with 99.75% accuracy for the wood class, GoogLeNet with 99.44% accuracy for the water class and VGGNet19 with 99.15% accuracy for the foliage class. Being so close to 100%, it can be emphasized the fact that those categories consist of a larger amount of data for training and testing as well as the possibility of over-fitting to happen. Another main observation is about the fact that both leather and water are two categories that obtain very good results overall across multiple models tested and that is due to the reflectance information that this dataset offers. In [23], the researchers make a comparison using their own convnet for training the network between FMD and the newly developed database, GMD, where it is emphasized that the considerable number of data for leather allows the system to isolate and distinguish between leather and fabric despite their many common characteristics. The same behavior is suggested by the data in Table 7 where the leather label has been identified better than the fabric label. Applying the same level of thought, if we are looking at other two labels: paper and foliage that share various common properties, the same behavior is inexistent as the results across all CNNs tried are very similar. Being a dataset that has only been released recently, it leaves room for improvement when it comes to solving the challenging task of recognizing materials.

Table 7 – GMD dataset results

| GMD | Fabric | Foliage | Glass | Leather | Metal | Paper | Plastic | Stone | Water | Wood |
|---|---|---|---|---|---|---|---|---|---|---|
| GoogLeNet | 78.17 | 97.26 | 85.52 | 95.96 | 84.04 | 90.23 | 77.61 | 96.7 | **99.44** | 96.91 |
| ResNet50 | 99.99 | 99.66 | 99.96 | **100** | 99.65 | 99.21 | 98.82 | 99.9 | 98.37 | 99.95 |
| ResNet101 | 99.92 | 99.5 | 99.86 | **100** | 99.67 | 99.07 | 99.57 | 99.69 | 99.01 | 99.75 |
| ResNet152 | 99.71 | 99.71 | 99.83 | **100** | 99.41 | 99.73 | 99.23 | 99.68 | 98.8 | 99.91 |
| VGG_CNN_S | 83.87 | 96.65 | 90.91 | **97.78** | 85.59 | 97.07 | 60.76 | 96.62 | 97.44 | 97.19 |
| VGG_CNN_M | 88.52 | 96.76 | 89.48 | 94.28 | 75.34 | 95.86 | 50.01 | 95.64 | **96.77** | 95.95 |
| VGG_CNN_F | 81.37 | 97.37 | 86.37 | 96.98 | 78.72 | 92.29 | 76.18 | 94.57 | **98.93** | 95.04 |
| VGGNet16 | 91.15 | 98.71 | 95.24 | 97.72 | 93.06 | 98.43 | 90.14 | 94.67 | 99.52 | **99.75** |
| VGGNet19 | 88.76 | **99.15** | 92.06 | 97.68 | 88.79 | 97.94 | 86.05 | 95.05 | 99.08 | 98.55 |

One new thing that my project focused on was understanding which architecture achieves the best results in the least amount of time. As networks tend to become deeper and more complex as well as more and more images being collected for the creation of larger datasets, the computational speed increases at a slow rate for the tasks that are needed to be completed in a timely manner. For instance, due to the fact that training a new network from scratch would potentially take weeks up to months to be completed, the system includes a pre-trained network on the ImageNet dataset that has around 1.2 million images. This allowed for time to be spent analyzing the different architectures and how they can be adapted to the Caffe framework in order for the experimental tests to begin.

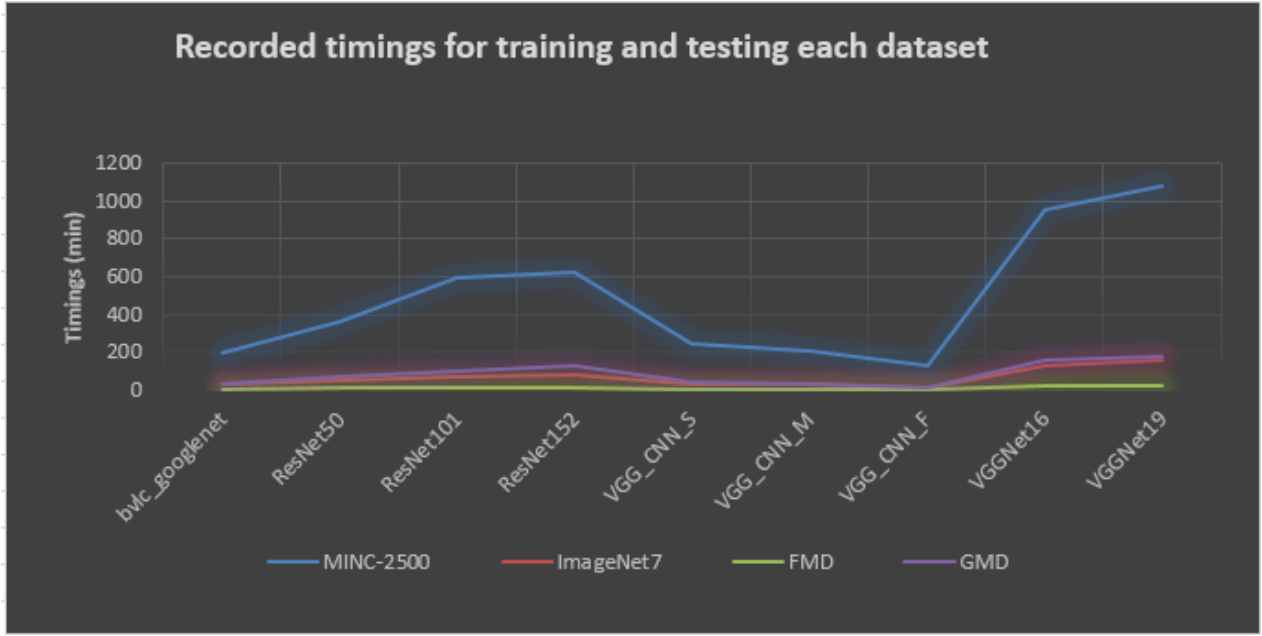

Figure 14 – Graph of recorded tests timings

As it is shown in Figure 14, the larger the dataset is, the more time running different batches of tests take. Looking at the architectures, the two types of VGGNet with 16 and 19 layers as well as the three types of ResNet with 50, 101 and 152 layers take the longest in order to process all the information found in an image compared to the remaining models tested. It is suggested by the results that the deeper and more complex the network is, the longer it takes the algorithm to identify all the interest points from the patches extracted and go through all the images as well as computing the image rankings and displaying the results. The graph displays the time recorded in minutes for all tests run for each individual model. In order to get a better understanding of the timings for each dataset, the data in Table 8 will be evaluated. As it was expected, the MINC2500 dataset took the longest to finish all combinations of tests that were run on the 23 total categories in 73 hours and 11 minutes. The recently published GMD dataset [23], successfully finished running all the tests in around 13 hours and 1 minute, generating a good overall timing for the dataset. The remaining datasets achieve the best timings out of all the tried datasets where tests performed on ImageNet7 complete in just 4 hours and 38 minutes and tests performed on FMD complete in one hour and 49 minutes. The timings reflect and are direct proportional to each dataset size. The main contribution from collecting and analyzing this data is that the number of images that are fed as the system's inputs represent an important factor that affects the pipeline accordingly.

Table 8 – Timings for all experimental test runs

| CNN | MINC-2500 | ImageNet7 | FMD | GMD |
|---|---|---|---|---|
| GoogLeNet | 3h 15m | 35m | 5m | 36m |
| ResNet50 | 6h 5m | 58m | 11m | 1h 18m |
| ResNet101 | 10h | 1h 18m | 12m | 1h 38m |
| ResNet152 | 10h 20m | 1h 20m | 13m | 2h 15m |
| VGG_CNN_S | 4h 7m | 35m | 9m | 41m |
| VGG_CNN_M | 3h 25m | 35m | 9m | 33m |
| VGG_CNN_F | 2h 7m | 20m | 6m | 20m |
| VGGNet16 | 15h 50m | 2h 13m | 22m | 2h 37m |
| VGGNet19 | 18h 2m | 2h 44m | 22m | 3h 3m |
| TOTAL | 73h 11m | 4h 38m | 1h 49m | 13h 1m |

Looking from the CNN perspective, the fastest model that has run across all four datasets is VGG_CNN_F completing training and testing the network in just two hours and fifty-three minutes, followed by GoogLeNet that finishes in around four hours and thirty-one minutes. These two architectures have the same structure for the first convolutional layer having 64 parameters and 7 convolutional kernels. The only difference between GoogLeNet and VGG_CNN_F is that the first has a stride of two pixels, while the later has a stride of four pixels. As it is observed in Table 9, the deeper the architectures have their network and the more complex structure, the longer it takes to extract all the necessary features to be passed to the classifier. On the contrary, the worst performing architecture in regard to speed computation is the VGGNet16 model that takes twenty-one hours and two minutes to finish. Since the total is represented by all the tests run across all the datasets, the architectures are analyzed from an objective point of view.

Table 9 – CNN timings

| Method | Timing hours |
|---|---|
| GoogLeNet | 4h 31m |
| ResNet50 | 8h 32m |
| ResNet101 | 13h 8m |
| ResNet152 | 14h 8m |
| VGG_CNN_S | 5h 32m |
| VGG_CNN_M | 4h 42m |
| VGG_CNN_F | 2h 53m |
| VGGNet16 | 21h 2m |
| VGGNet19 | 24h 11m |

Training and testing state-of-the-art CNNs for the task of material classification with images from the real-world, offers very promising results and interesting directions and approaches to improve. Understanding the requirements of a good network for image classification starts with trying the latest techniques and methods and adapting them to the newest tasks.

# Chapter 4 Project planning

As it can be outlined in the above-mentioned chapters, this research project has provided several challenging tasks such as becoming familiar with the diverse topics related to my chosen subject in order to make sure that the product deliverable is at a high standard, and estimating the length of tasks throughout the year. This section outlines how the methodology has been used in the planning of the project and gives a description of any resources and software used.

## 4.1 Planning

The learning process has started early during the summer period 2016 and responsibilities such as setting up the system, running initial tests to understand the algorithms and getting background knowledge have been the central focus. Running a few tests allowed me to understand how much time is needed for the tests estimation. Improving my existent awareness of the field, I have used available resources and did my literature survey on the topic. Starting with three fundamental books [39, 40, 41] for understanding the mathematics for machine learning and applications of various classification algorithms, the first month was spent on understanding the complex process of image formation and classification using feature detection. The next three weeks have been focused on a series of practicals authored by Vedaldi and Zisserman [25] from University of Oxford, Visual Geometry Group as they cover related topics to my research project through a series of exercises in MATLAB. Observing the way the network interacts with a wide range of Convolutional Neural Networks architectures to detect, recognize or classify categories, I was able to extract information and code about how a classifier works on a particular dataset. Due to the current project management being implemented effectively, I was able to use Coursera as an online resource to increase my knowledge on deep neural networks and classifiers and enrolled within two modules: Neural Networks for Machine Learning [42] created and delivered by Geoffrey Hinton and Machine Learning [43] created by Andrew Ng from Stanford University. The two modules gave valuable information on how the network's weights are learned and transmitted within the network as well as understanding the mathematics and code behind diverse approaches used on convolutional layers. As the literature has been completed, the planning for the remaining two terms involved setting up the environment and starting the test. These were constantly updated in the logbook and their estimates updated to reflect the current state of the work at all stages of the project.

## 4.2 Methodology

The Waterfall Methodology has been used since the project presents various characteristics of such a methodology. Having clear stated goals broken down into a series of phases from the start of the project is the core of turning research work into reliable deliverables as well as producing constant results. A Waterfall Methodology consists of milestones, a consistent process and estimations. Milestones are the predefined steps or topics that mark together the completion of the project. In the context of this research work, they are: Summer Preparation, System Creation and Tests. These represent the foundation that has been divided in the design stage, integration stage, testing stage and the evaluation stage for each Convolutional Neural Network. The estimates used throughout the project outline the scope and progress made. Having such an important role in this type of methodology, they were made at the beginning of each milestone and they were updated as tasks would be completed.

The Technical Documentation and Results chapter highlights the solid knowledge needed for the chosen research topic. Getting familiar with the field implied planning the summer work to be spent on learning about the state-of-the-art techniques and architectures in Computer Vision as well as understanding how to find suitable research papers that contain applicable information to material classification. Using the suggestions recommended by my supervisors, I learned how to adapt to change and take advantage of the available resources to prepare to conduct reliable data collection that will result in the expected evaluation. At the end of the autumn term, I have found a new material dataset that was recently published that offered a promising direction to how materials can be identified in an image. Due to the concluded results on the new database, the project has been modified in order for the new dataset to be added to the current structure. By recalculating the estimates as well as introducing new tasks and their deadlines, the new collection of data has been set up and adapted.

Managing effectively the project involved reviewing and prioritizing the most important duties for the benefit of solving the challenging task. Writing entries in a weekly logbook supported the early identification of risks and planning in advance to minimalize them. Learning to be proactive has proved to be an asset skill. Having a section like reflection is essential as it captures the review of the work completed each week and focuses on what needs to be done next. Risk management has been implemented from the start of the project to make sure that their occurrence is avoided and that, if a risk is encountered, the liability will be analyzed and handled. This is favorable in dealing with risks and reflecting on the progress made so far.

The chosen methodology has proven to be suitable for the task and lead to the achievement of producing a good evaluation of each component of the pipeline. The overall performance of the system improves earlier work done on material recognition and classification achieving 92.56% accuracy on MINC2500 dataset, 79.08% accuracy on ImageNet7 dataset, 91.86% accuracy on FMD dataset and 99.75% accuracy for the wood category from the GMD dataset.

## 4.3 Tools and Software

To keep track of all the tasks and the efforts put in, a logbook and a Gantt Chart have been used and have been weekly updated. The logbook was a beneficial tool that aided keeping a constant good overall organization and awareness of the plan for each week, the content and reflection on the methods and outcomes reached. At the end of each planned week, the Gantt Chart would be updated with the latest estimations and percentages of work completed. By keeping the momentum throughout the duration of the project at a very high standard in terms of project management, it leads to successfully delivering the experimental test results. The Gantt Chart tool has been very useful for scheduling tasks and improving the project's planning. Being amended at the end of each week, it reflected the exact stage that I was working on and allowed to be aware of the estimates at all times throughout the duration of the project.

The software used was MATLAB. This is a programming environment used predominantly in Computer Vision research work. Having prebuilt functions and programming languages for the simulation of a neural network, it leads to faster completion of the setting up the environment stage. One important tool that is going to be extensively used for obtaining the results is the plotting algorithm. This will extract the score maps from the classifier for each network model and output it to the user. The code has been downloaded from Kalliatakis et al [44] and adapted for the task of material classification. The MatDeepRep function extracted is a deep representation tool in MATLAB that is mainly used for image classification. It uses pre-trained models to easily train and test a wide range of CNN architectures, datasets and SVMs. In order to be able to use this function, some minor adaptations had to be made to run on the set environment as well as a new function called "batch_of_tests.m" was created. This method will be used to call the "matdeeprep.m" method. To run a test, a fixed number of parameters need to be used such as: the name of the CNN architecture, the name of the dataset that is going to be run on, the name of the category followed by the name of the test that will be used when saving the outputs. The results consist of four figures in total: two for the training phase and two for the testing phase. For instance, for the training, a figure containing the top 36 best-ranked images will be saved as well as a graph of the mean AP computed on the score map.

## Chapter 5 Conclusion

Due to the variability of the materials, recognizing materials using supervised learning is a very challenging task that has received lots of attention in the last decades. This project targets specifically material classification and presents an accurate and observational evaluation of nine distinct CNN models on four varied datasets. Since segmentation and image understanding are some of the fundamental challenges computer vision systems attempt to tackle, this project took a further look at approaches like patch segmentation and transfer learning and how they affect the way the features are learned by the network at different layers.

Through the experimental tests, the real-world scene understanding has been examined by looking at the contextual modelling between the diversified components of the created system. The pipeline consists of the training and testing sets that are fed as inputs to the pre-trained network on the ImageNet dataset. The network will then extract the features in a feature vector that is fed to the classifier which will compute the score map. The mean average precision will rank each image in the dataset and output the results.

The results gathered establish that the state-of-the-art accuracy of up to ~92.56% is achieved using a deep neural network on the MINC2500 dataset when patch segmentation and transfer learning are used as methods. FMD manages to obtain very equivalent results with just a -0.7% drop in accuracy, being the second top model that manages to correctly identify the feature labels. The current system improves the overall results for the ImageNet7 database by at least ~4% achieving a 79.08% performance. The extensive testing on the GMD dataset provides very good results with the top three categories being wood – 99.75%, water – 99.44% and foliage – 99.15%. All results obtained from training the network for material classification has shown an improvement in the overall performance when a simple linear support vector machine has been combined with nine distinct CNNs and four material databases. To make sure that the outcomes outline the full complexity of the tests, all categories from each dataset have been run. What it is brought new to the scene is the analysis of the how fast the architectures are as well as observing what factors can influence the system. The number of images offered by each database affects the amount of time the architectures take to evaluate all architectures differently in regard to the size of each dataset. The overall timings differ from around one hour and 49 minutes for the FMD dataset to finish all tests up to 73 hours and 11 minutes for MINC2500 dataset to successfully complete a batch of tests.

## 5.1 Outlook

In addition to answering a few questions about image recognition and material classification, the work completed has opened countless opportunities for future research topics and experiments. This section will outline a few of the outlook considered.

Comparison of the nine CNN methods showed that patch segmentation achieves very superior results, which opens directions for other techniques such as full semantic segmentation combined with transfer learning to be investigated on the state-of-the-art results. Since patch segmentation is such promising approach, it can be optimized and improved by introducing a range of pre-processing steps such as reducing the noise or adapting the normalization that can lead to a better performance of the technique. In the context presented above, transfer learning has shown to have improved the results compared to other previous work that has been discussed in the first section. Currently, the algorithm extracts information from the layer before the last layer that holds information such as shading and reflectance. An interesting further step would be to access the information from the last convolutional layer that would provide information about the texture of the material and see how the results and the overall system is affected.

The current pipeline has a fixed linear SVM. Further work would be orientated toward adapting various types of classifiers to understand which combination of neural network would be the most accurate for analyzing material categories. By comparing multiple distinct types of classifier such as multiclass or with polynomial kernels, the network can improve the overall performance.

# Appendix

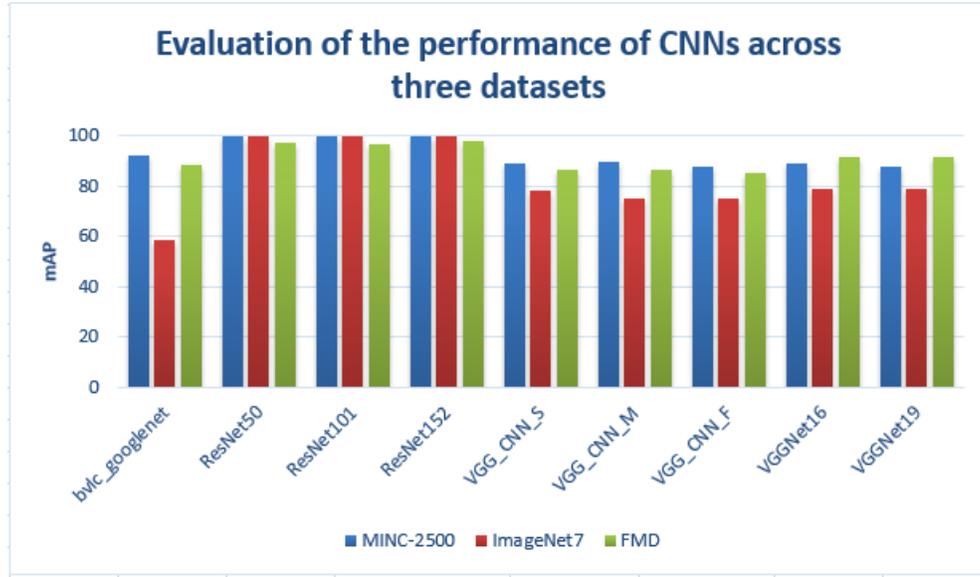

Figure 15 – Graph of MINC2500, ImageNet7 and FMD performances

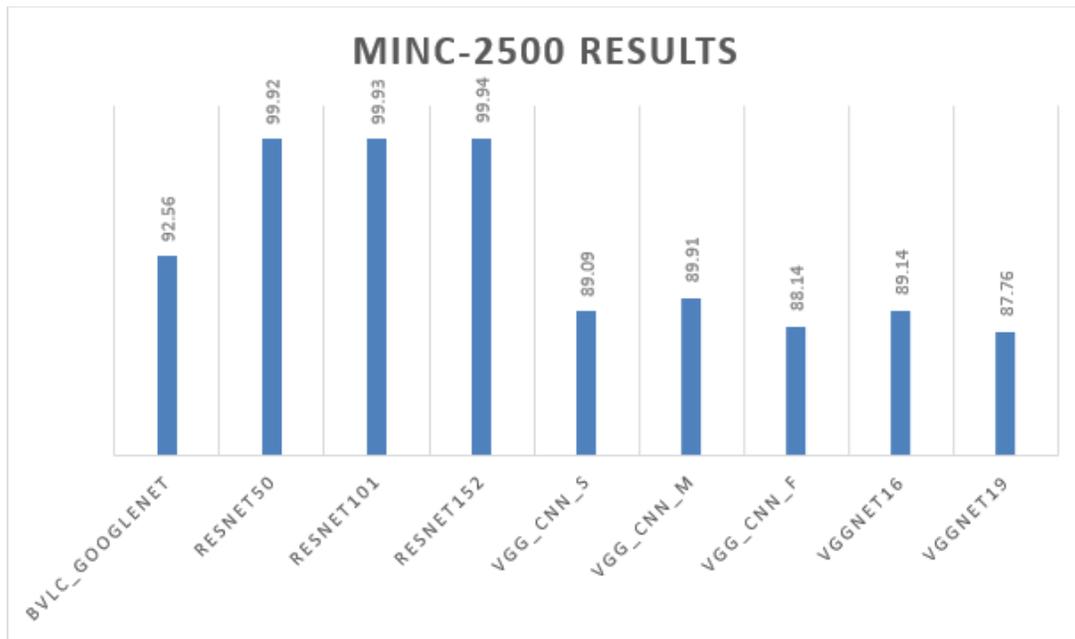

Figure 16 – MINC2500 results

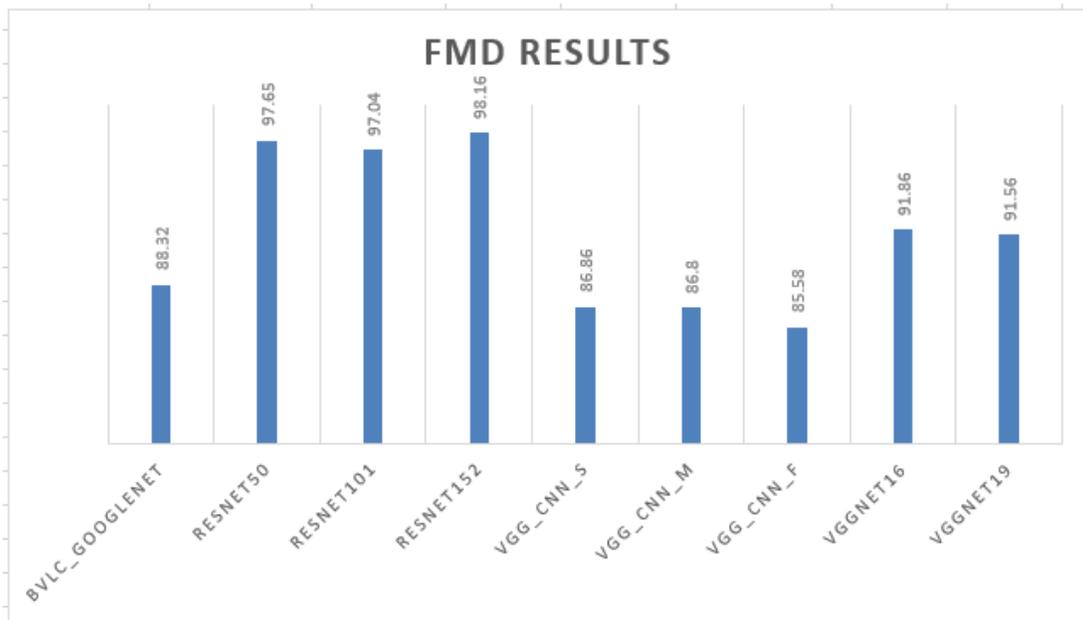

Figure 17 – FMD results

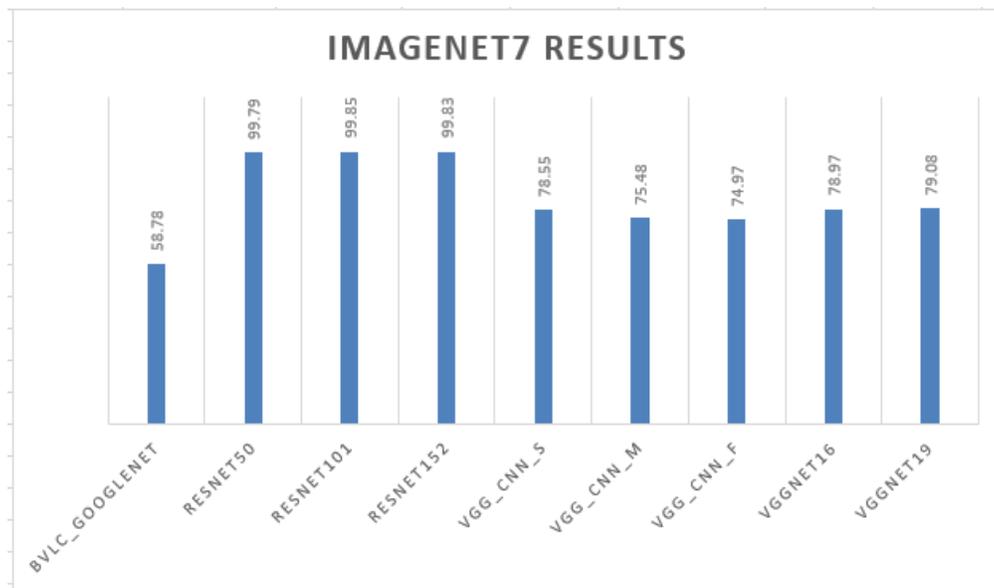

Figure 18 – ImageNet7 results

Table 10 – Timings expressed in minutes as extracted from the data collection

| CNN | MINC-2500 | ImageNet7 | FMD | GMD |
|---|---|---|---|---|
| GoogLeNet | 195 | 35 | 5 | 36 |
| ResNet50 | 365 | 58 | 11 | 78 |
| ResNet101 | 597 | 78 | 12 | 98 |
| ResNet152 | 620 | 80 | 13 | 135 |
| VGG_CNN_S | 247 | 35 | 9 | 41 |
| VGG_CNN_M | 205 | 35 | 9 | 33 |
| VGG_CNN_F | 127 | 20 | 6 | 20 |
| VGGNet16 | 949 | 133 | 22 | 157 |
| VGGNet19 | 1082 | 164 | 22 | 183 |
| TOTAL | 4387 | 638 | 109 | 781 |